\documentclass[letterpaper, 10 pt, conference]{ieeeconf}  %IEEE字体
\IEEEoverridecommandlockouts   % 允许使用 \thanks
\usepackage{amsfonts}
\usepackage{amsmath}
\usepackage{algorithm} %算法
\usepackage{algpseudocode} %算法伪代码
\usepackage{bm} 
\usepackage{float}
\usepackage{graphicx}
\usepackage{subfigure}
\usepackage{cite}
\usepackage[colorlinks=true, allcolors=blue]{hyperref}
\usepackage[utf8]{inputenc}
\usepackage[T1]{fontenc}
\usepackage[capitalise]{cleveref}
 \usepackage{balance}

\begin{document}

%------------标题------------
\title{
Like Playing a Video Game: Spatial-Temporal Optimization of Foot Trajectories for Controlled Football Kicking in Bipedal Robots
}
%------------作者------------
\author{Wanyue Li$^{*}$, Ji Ma$^{*}$, Minghao Lu and Peng Lu$^{\dagger}$% <-this % stops a space
\thanks{$^{*}$ Equal Contribution.}
\thanks{$^{\dagger}$Corresponding author: \url{lupeng@hku.hk} }
\thanks{The authors are with the Adaptive Robotic Controls Lab (ArcLab), Department of Mechanical Engineering, The University of Hong Kong 710075, Hong Kong. mail: {\tt\footnotesize liwy1024@connect.hku.hk; minghao0@connect.hku.hk
maji@connect.hku.hk; lupeng@hku.hk}}%
\thanks{Project materials: \url{https://arclab-hku.github.io/STOFT/}.}
}
 
% === 临时调整第一页顶部空白 ===
\maketitle
\pagestyle{empty}  % no page number for the second and the later pages
\thispagestyle{empty} % no page number for the first page

%%%%%%%%%%%%%%%%%%%%%%%%%%%%%%%%%%%%%%%%%%%%%%%%%%%%%%%%%%%%%%%%%%%%%%%%%%%%%%%%
%------------摘要------------
%%%%%%%%%%%%%%%%%%%%%%%%%%%%%%%%%%%%%%%%%%%%%%%%%%%%%%%%%%%%%%%%%%%%%%%%%%%%%%%%
\begin{abstract}
Humanoid robot soccer presents several challenges, particularly in maintaining system stability during aggressive kicking motions while achieving precise ball trajectory control.
Current solutions, whether traditional position-based control methods or reinforcement learning (RL) approaches, exhibit significant limitations. 
Model predictive control (MPC) is a prevalent approach for ordinary quadruped and biped robots.  While MPC has demonstrated advantages in legged robots, existing studies often oversimplify the leg swing progress, relying merely on simple trajectory interpolation methods. This severely constrains the foot's environmental interaction capability, hindering tasks such as ball kicking.
This study innovatively adapts the spatial-temporal trajectory planning method, which has been successful in drone applications, to bipedal robotic systems. 
The proposed approach autonomously generates foot trajectories that satisfy constraints on target kicking position, velocity, and acceleration while simultaneously optimizing swing phase duration. Experimental results demonstrate that the optimized trajectories closely mimic human kicking behavior, featuring a backswing motion. Simulation and hardware experiments confirm the algorithm's efficiency, with trajectory planning times under 1 ms, and its reliability, achieving nearly 100 \% task completion accuracy when the soccer goal is within the range of -90° to 90°.

% 这里要看足球相关论文，给出几个足球相关的衡量标准

\end{abstract}

% \begin{IEEEkeywords}
% Humanoid and Bipedal Locomotion, 
% Optimization and Optimal Control, 
% Legged Robots
% \end{IEEEkeywords}
 %%%%%%%%%%%%%%%%%%%%%%%%%%%%%%%%%%%%%%%%%%%%%%%%%%%%%%%%%%%%%%%%%%%%%%%%%%%%%%%%
 %------------介绍------------
%%%%%%%%%%%%%%%%%%%%%%%%%%%%%%%%%%%%%%%%%%%%%%%%%%%%%%%%%%%%%%%%%%%%%%%%%%%%%%%%

\section{INTRODUCTION}

%------------定妆照图片------------
\begin{figure}[]
\centering
\includegraphics[width=0.98\linewidth]{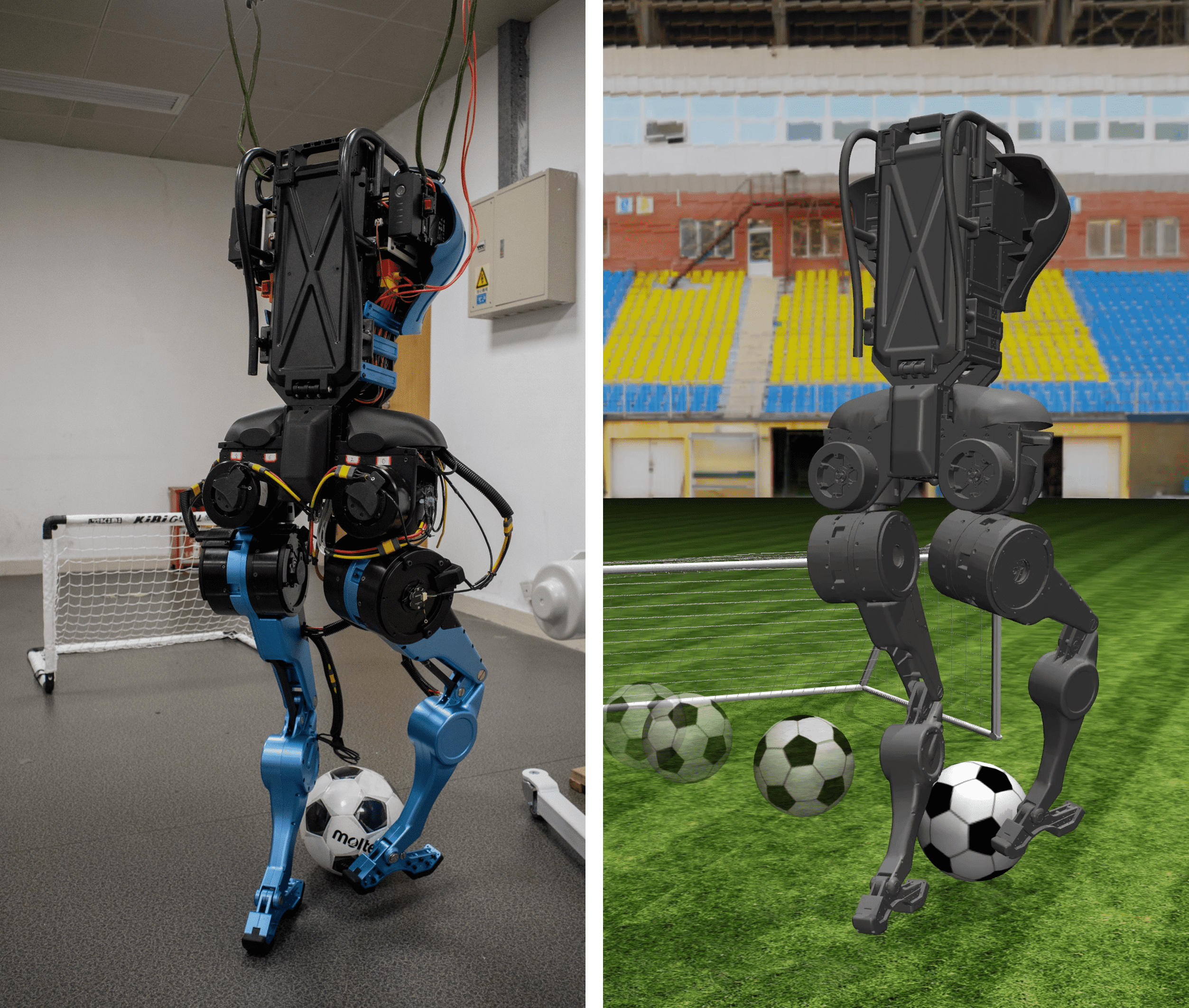}
% \caption{The PEARL biped robot has 5 degrees of freedom (DoFs) per leg, weighs 40 kg, and stands 1.5 meters tall.}    
\caption{
Real-world and simulated kicking experiments with the PEARL biped robot. The robot has 5 degrees of freedom per leg, weighs 40~kg, and stands 1.5~m tall.
}
\label{fig:robot}
\end{figure}

\begin{figure*}[t]
\centering
\includegraphics[width=\linewidth]{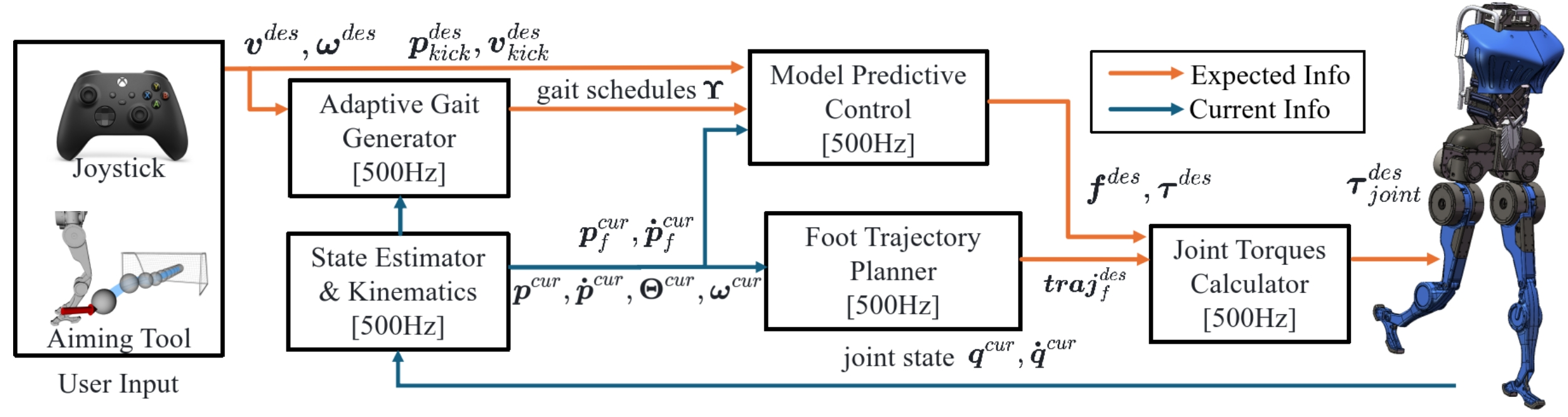}
\caption{
Control Framework: The robot operator provides control commands, including the desired robot velocity \((\boldsymbol{v}^{des}, \boldsymbol{\omega}^{des})\) and the kicking foot target \((\boldsymbol{p}_{kick}^{des}, \boldsymbol{v}_{kick}^{des})\).  
The state estimator calculates the current system states \((\boldsymbol{p}^{cur}, \boldsymbol{\dot{p}}^{cur}, \boldsymbol{\Theta}^{cur}, \boldsymbol{\omega}^{cur})\) and foot states \((\boldsymbol{p}^{cur}_{f}, \boldsymbol{\dot{p}}^{cur}_{f})\). The gait generator then produces an adaptive gait schedule \(\boldsymbol{\Upsilon}\).  
Using the control commands, current states, and gait schedule, the MPC module solves an OCP to compute the optimal GRFs and GRTs \((\boldsymbol{f}^{des}, \boldsymbol{\tau}^{des})\). Simultaneously, the foot planner generates a reference trajectory \(\boldsymbol{\phi}(t)\) for either executing a kick or regular walking.  
Finally, based on the desired reference trajectory, GRFs, and GRTs, the joint torques \(\boldsymbol{\tau}^{des}_{joint}\) are computed for precise actuation.
}
\label{fig:frame}
\end{figure*}
Football is one of the most popular sports worldwide, capturing the interest of billions. Given their human-like form, the application of humanoid robots in football naturally attracts significant interest. Since the inaugural RoboCup in 1997 \cite{kitano1997robocup}, the competition has served as a driving force for research in robot soccer, leading to advancements in areas such as visual perception and stable motion control.
However, despite significant progress in robotic football, achieving precise passing and shooting remains a major challenge that requires further investigation.

Most previous approaches have focused primarily on executing kicking motions rather than precisely controlling the placement of the shot. For instance, humanoid robots in RoboCup\cite{friedmann2008versatile},\cite{behnke2008hierarchical},\cite{carlos2008modular} mainly rely on predefined motions, while \cite{jouandeau2014optimization} employed parameterized multi-phase actions in simulation to extend shooting range.
In recent years, some studies have begun to emphasize precise passing, mainly through reinforcement learning (RL). For example, \cite{ji2022hierarchical} demonstrated that a quadruped robot could use RL to perform single-leg passing tasks from a static position, \cite{ji2023dribblebot} showcased its capability for remote-controlled dribbling. Similarly, \cite{liu2022motor}, \cite{haarnoja2024learning} developed a highly realistic physical system that closely matches its simulation counterpart, enabling small-scale humanoid robots to autonomously dribble, kick, and shoot in 1v1 soccer scenarios with the aid of motion capture systems.
However, RL-based methods are often tailored to specific robots and environments, limiting their ability to generalize to real-world scenarios. Furthermore, RL policies typically operate as a “black box,” making their decision-making process challenging to interpret and control. As a result, there remains a lack of humanoid robot systems capable of autonomously planning and executing precise shots based on human-specified targets. Our work aims to 
bridge this gap.

% \clearpage % 确保第一页结束，第二页开始用默认边距
Model Predictive Control (MPC) has become increasingly popular in legged robot control, achieving remarkable success in legged robot locomotion.
Convex MPC based on a single rigid body (SRB) model, the MIT Cheetah 3 can reach speeds up to 3.7 m/s \cite{di2018dynamic} \cite{kim2019highly}.  
Perceptive Locomotion through Nonlinear MPC \cite{grandia2023perceptive} enables quadrupedal robots to traverse complex terrains stably. Contact-Implicit MPC \cite{kim2024contact} eliminates the need for predefined contact sequences, allowing quadrupedal robots to achieve stable bipedal standing.
These advancements have also influenced bipedal locomotion.  
\cite{garcia2021mpc}\cite{li2021force} adapted MPC strategies for bipedal robots with line feet and explored MPC modeling for bipedal locomotion. CDM-MPC considers that the centroidal dynamics model can realize dynamic jumping\cite{he2024cdm}.
However, swing leg control has often been overlooked, and a simple Bézier curve is typically used. It primarily ensures foot placement at the designed touchdown time, treating it as a terminal constraint rather than tracking the entire trajectory \cite{daneshmand2021variable}. 
This highlights the underexplored potential of swing leg control, warranting further investigation.

During human kicking motions, movement is naturally adjusted based on the target distance: for short-range passes, a slight leg swing suffices, whereas long-range passes or shots require a more pronounced backswing. The backswing serves to extend both the acceleration distance and duration, allowing the foot to reach a higher velocity at the moment of impact and achieve the desired motion state at the target location. Fundamentally, this process involves the simultaneous optimization of trajectory shape and timing parameters, a concept known as spatial-temporal trajectory planning \cite{wang2022geometrically}.

This approach has been widely applied in drone trajectory optimization. \cite{zhou2021ego} and \cite{zhou2022swarm} demonstrated coordinated outdoor flight, obstacle avoidance, and inter-robot collision avoidance in multi-drone systems. \cite{wang2024implicit} ensured continuous, collision-free motion for drones navigating complex environments with non-convex geometries. In real-world tests, SUPER achieved autonomous flights at speeds exceeding 20 meters per second \cite{ren2025safety}.

Inspired by human kicking mechanics and leveraging spatial-temporal trajectory planning \cite{lu2024fapp}, this study introduces the Spatial-Temporal Optimization of Foot Trajectory (STOFT) Planner, a novel approach designed to optimize the kicking trajectory of humanoid robots.
% 在人类踢球过程中，踢球动作会根据目标距离进行调整：近距离传球时，仅需小幅度抬腿击球，而长距离传球或射门则伴随大幅度的后摆动作。后摆动作的作用在于延长加速距离，增加加速时间，使脚在击球瞬间达到更高的速度，并在目标点处实现预期的运动状态。本质上，这是一个同时优化轨迹形状与时间参数的过程，即 spatial-temporal trajectory planning \cite{wang2022geometrically}。
% 该方法已广泛应用于 无人机轨迹优化 领域。
% \cite{zhou2021ego}\cite{zhou2022swarm}实现了多架无人机户外协同飞行和, obstacle avoidance, and inter-robot collision avoidance.
% \cite{wang2024implicit} ensuring continuous collision-free motion for non-convex geometries in complex environments.
% \cite{ren2025safety} In real-world tests, SUPER achieved autonomous flights at speeds exceeding 20 meters per second.
% 本研究受人类踢球动作的启发，结合spatial-temporal trajectory planning方法\cite{lu2024fapp}，提出了一种新颖的类人机器人射门轨迹规划器 Humanoid Foot Spatial-Temporal Trajectory (HFSTT) Planner.

%本篇论文的贡献就是:
The main contributions of this paper include:
\begin{itemize}
\item A comprehensive system architecture for dynamic and precise shooting, integrating vision-based soccer detection and localization, visual aiming, robot state estimation, MPC-based balance control, spatial-temporal foot trajectory planning, and adaptive gait generation for uncertain swing phases

\item Proposed the STOFT Planner, which performs spatial-temporal optimization to execute precise kicking tasks while adhering to dynamic constraints and avoiding self-collision. This approach generates natural backswing motions, closely mimicking human kicking behavior.

\item Developed the humanoid robot PEARL for real-world testing. Experimental results show that the system can perform precise shooting football, with the planner completing a single computation in under 1 ms.

% Furthermore, the robot exhibits strong disturbance rejection and the ability to autonomously avoid obstacles in complex environments, enhancing its overall adaptability and stability.

\end{itemize}
 
%  In the simulation, the Yat-sen Lion can transform its trunk while trotting at a speed of 2.8 m/s or an angular velocity of 6 rad/s. In the hardware experiments, it can achieve 0.5 m/s and 0.5 rad/s. Due to the actuated spine, it can lift the equipment to 0.54 m to investigate, pass through a narrow tunnel, and climbs a slope or cross uneven ground while keeping the middle part of the trunk parallel to the horizontal plane.
 
%这次主要工作在于对力控脊椎四足的控制，未来的工作，给脊椎，和腿部摆动配合起来，能量

% 本文的其余部分安排如下。在第二部分，我们讨论了相关的工作。在第三节中，展示中山狮子的机械设计。在第四节中，阐述MPC和浮动基动力学。在第五部分，我们进行了现场测试，用来检测算法的可靠性。最后，第六部分对本文进行总结。
The rest of this paper is arranged as follows. Section II presents the system pipeline. Section III introduces the STOFT planner. In Section IV, experimental results verify the reliability of the proposed method. Finally, Section V concludes this paper.
 
   %------------框架图片------------

\section{SYSTEM FRAMEWORK}

 %------------刚体运动学树图片------------
\begin{figure*}[ht]
    \centering
    \subfigure[]{
        \begin{minipage}[b]{0.75\linewidth}
            \includegraphics[scale=0.48]{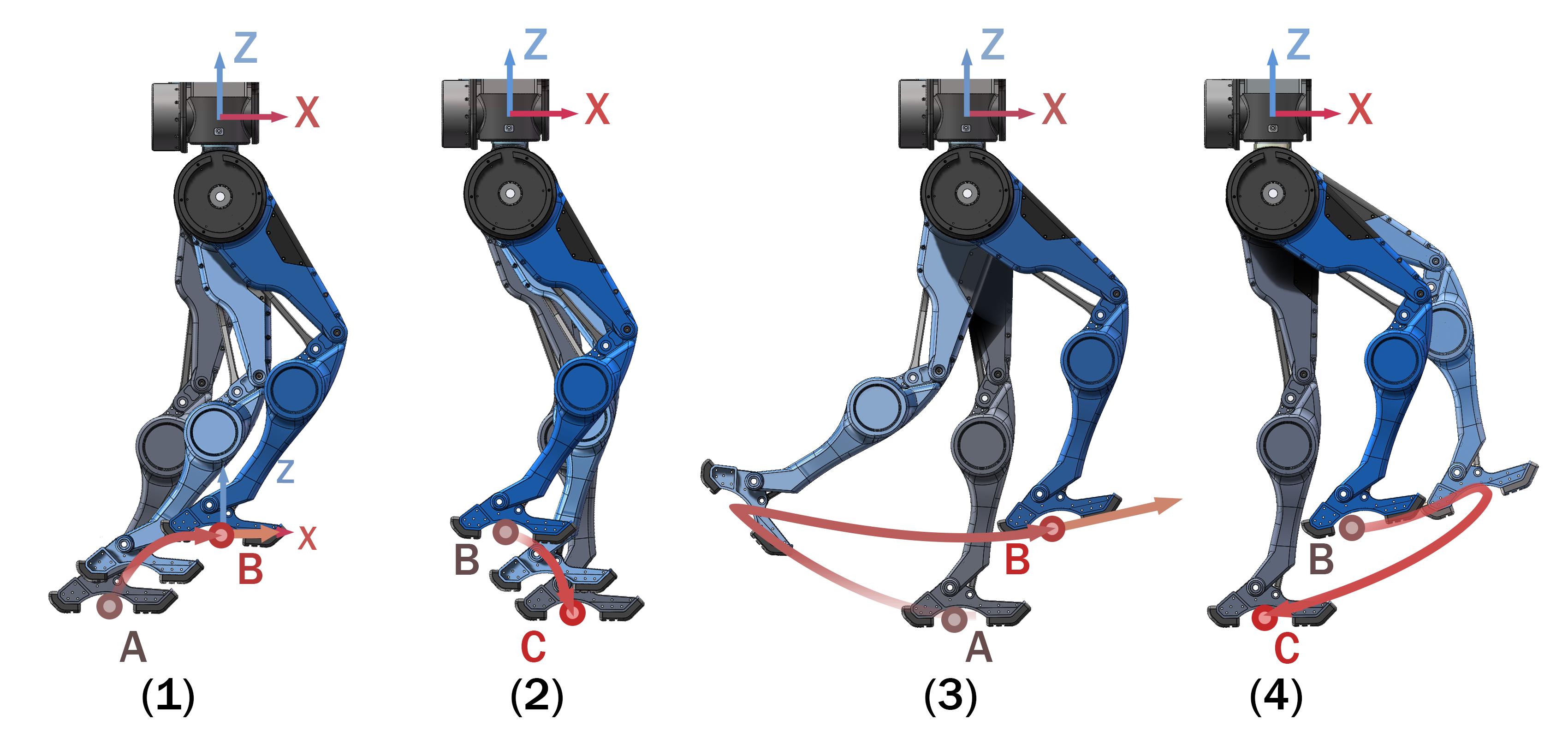}
        \end{minipage}
        \label{fig:rigidBody}
    }\subfigure[]{
        \begin{minipage}[b]{0.25\linewidth}
            \includegraphics[scale=0.48]{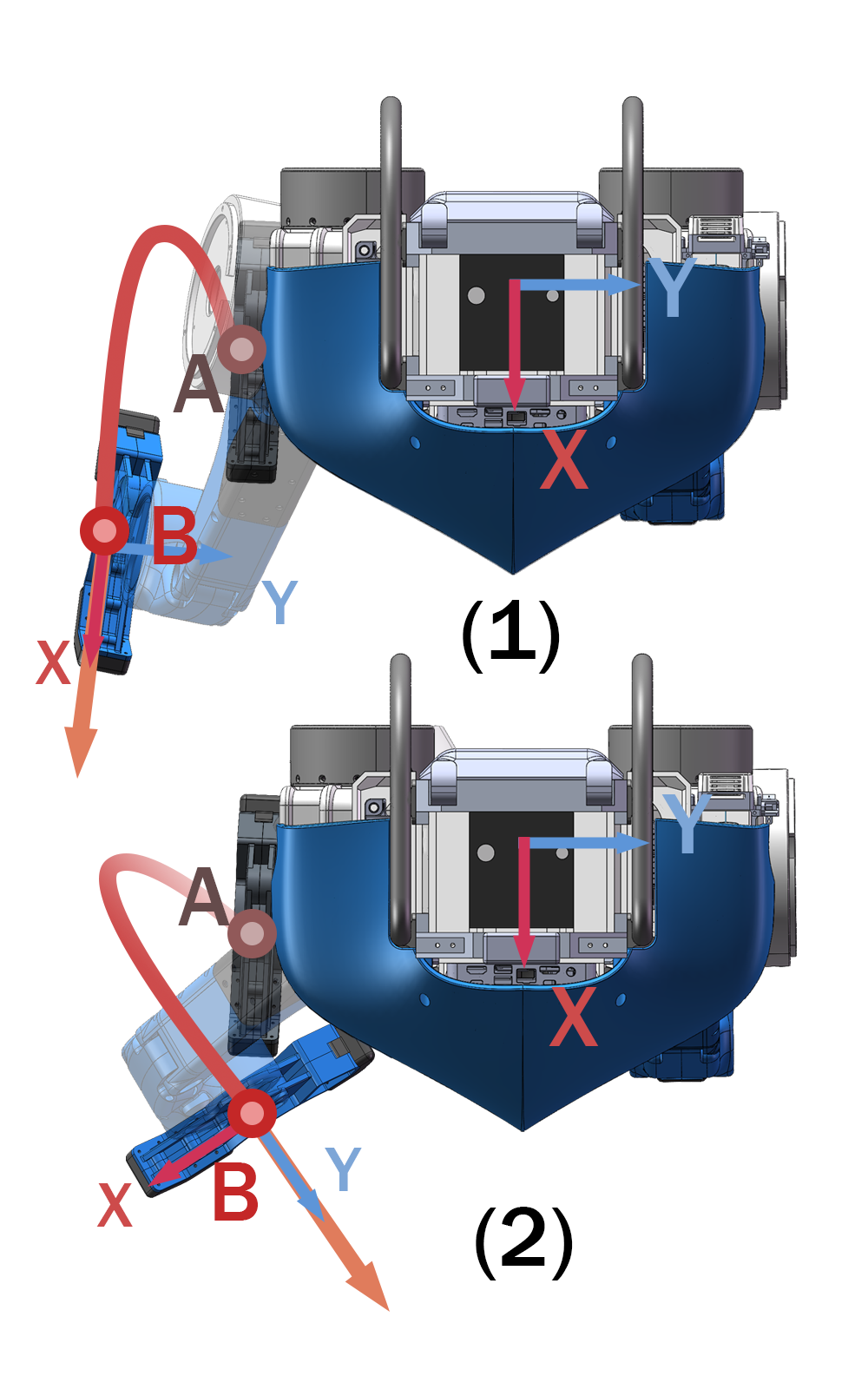}
        \end{minipage}
        \label{fig:connect}
    }
    \caption{
        (a) Side view showing the foot trajectory. (b) Top-down view illustrating the foot orientation during the kicking task. 
        Points A, B, and C represent the initial position, intermediate target, and desired foothold, respectively. 
        The orange arrow indicates the kicking velocity, and the red trajectory is optimized to satisfy position, velocity, and dynamic constraints at points A, B, and C. 
        In (a), (1) and (2) illustrate regular walking, while (3) and (4) demonstrate a high-velocity kick. 
        In (b), (1) shows a straight kick, and (2) depicts a side kick.
    } 
    \label{fig:kick}
\end{figure*}

   %------------质量集中模型图片------------

% The biped robot's complexity makes direct optimization computationally intensive and impractical for real-time use on embedded systems. To address this, we adopt the single rigid body (SRB) model, leveraging the fact that most of the robot's mass is concentrated in its main body.

% As shown in Fig. \ref{fig:robot}, 

The whole control framework is shown in Fig. \ref{fig:frame}. At each sampling interval, the MPC solves a finite-horizon Optimal Control Problem (OCP). It utilizes the current state estimated by the state estimator to track the reference trajectory. The OCP minimizes the cost function and computes the optimal sequences of ground reaction forces (GRF) and ground reaction torques (GRT). The control input for the first step of the computed sequence is then applied in the following sampling period.
During execution, the stance leg generates the required GRFs and GRTs, while the swing leg follows the desired trajectory produced by the STOFT planner. The joint torques are computed using the Jacobian matrix.

\subsection{Model Predictive Control}

As shown in Fig. $\ref{fig:robot}$, the leg masses are relatively light compared to the robot's body and can therefore be neglected. Consequently, the robot is modeled as a Single Rigid Body (SRB) \cite{li2021force}. Following the simplification proposed in \cite{di2018dynamic}, which assumes small roll and pitch angles, the dynamics of the SRB are expressed as follows:
\begin{equation}
\begin{gathered}
{}^w\dot{\boldsymbol{p}}={}^w \boldsymbol{v} \\
m {}^w\dot{\boldsymbol{v}}= {}^w \boldsymbol{F}  \\
\dot{\boldsymbol{\Theta}}=\boldsymbol{R}_z{} {}^{b}\boldsymbol{\omega} \\
\frac{d}{d t}({\boldsymbol{I}} \;{}^{b} \boldsymbol{\omega})={}^w_b \boldsymbol{R}^{T} \ {}^w \boldsymbol{M},  \\
\end{gathered}
\label{eq:srbd}
\end{equation}
where $\boldsymbol{\Theta} = \left[\phi, \theta, \psi\right]^{T} \in \mathbb{R}^{3}$ represents the Z-Y-X Euler angles, and $\boldsymbol{\omega} \in \mathbb{R}^{3}$ is the angular velocity of the robot's base frame. The vector $\boldsymbol{p} \in \mathbb{R}^{3}$ denotes the position of the center of mass (CoM), and $\boldsymbol{v}$ is the CoM's velocity. The scalar $m$ denotes the robot's mass, while $\boldsymbol{I}$ is the robot's inertia. The superscripts $w$ and $b$ refer to the world and body frames, respectively. The matrix $\boldsymbol{R}_{z}$ represents a rotation of the yaw angle $\psi$ about the Z-axis.
The net ground reaction forces ${}^w \boldsymbol{F}$ and torques ${}^w \boldsymbol{M}$ acting on the robot's CoM are defined as follows:

\begin{equation}
\begin{gathered}
{}^w \boldsymbol{F}=\sum_{l=1}^{n_{c}} {}^w \boldsymbol{f}_{i}+m \boldsymbol{g} \\
{}^w\boldsymbol{M}=\sum_{l=1}^{n_{c}}\left({}^w \boldsymbol{p}_{f,i}-{}^w \boldsymbol{p}\right) \times {}^w \boldsymbol{f}_{i}+ \boldsymbol{\tau}_i,\\
\end{gathered}
\end{equation}
where $\boldsymbol{p}_{f,i}$ is the position of the $i$-th leg. $n_c$ is the number of feet in stance phase and $\boldsymbol{g}$ is the gravitational acceleration.
$\boldsymbol{f}_{i}=[{f}_{x,i},{f}_{y,i},{f}_{z,i}] \in \mathbb{R}^{3}$ and $\boldsymbol{\tau}_i=[0,{\tau}_{y,i},{\tau}_{z,i}]\in \mathbb{R}^{3}$ represent the GRF and GRT of $i$-th contact point, respectively. Therefore, the system state is represented as:
\begin{equation}
\boldsymbol{x} =\left[\begin{array}{llll}%-------------公式1
\boldsymbol{\Theta}^{T} & \boldsymbol{p}^{T} & \boldsymbol{\omega}^{T} & {\boldsymbol{v}}^{T}
\end{array}\right]^{T}.
\end{equation}

Since the robot's feet cannot generate external moments around the x-axis, the system input is represented as the wrench exerted by both feet:
\begin{equation} 
\boldsymbol{u} = \left[ \begin{array}{c} {}^w \boldsymbol{f}_1 \ {}^w \boldsymbol{\tau}_1 \ {}^w \boldsymbol{f}_2 \ {}^w \boldsymbol{\tau}_2 \end{array} \right] \in \mathbb{R}^{12},
\end{equation}
where
${}^w \boldsymbol{f}_{i}=[{f}_{x,i},{f}_{y,i},{f}_{z,i}] \in \mathbb{R}^{3}$ and ${}^w \boldsymbol{\tau}_i=[0,{\tau}_{y,i},{\tau}_{z,i}]\in \mathbb{R}^{3}$ represent the GRF and GRT of $i$-th contact point, respectively. The $i$-th leg wrench is denoted by $\boldsymbol{u}_i = \left[ \begin{array}{c} {}^w \boldsymbol{f}_i \ {}^w \boldsymbol{\tau}_i \end{array} \right] \in \mathbb{R}^{6}$.
The constraints on the system include friction cone, force limits, and torque bounds:
\begin{equation}
\begin{aligned}
  - \mu f_{z,i} &\leq f_{x,i} \leq \mu f_{z,i} \\
 - \mu f_{z,i} &\leq f_{y,i} \leq \mu f_{z,i} \\
 0 \leq f_{z,i}& \leq f_{\max} \\
 |\tau_{y,i}|& \leq \tau_{\max} \\
 |\tau_{z,i}| &\leq \mu_{rot}  \ f_{z,i}
\label{eq:constrain},
\end{aligned}
\end{equation}
where $\mu$ is the translational friction coefficient, and $\mu_{rot}$ is the rotational friction coefficient. $f_{\max}$ and $\tau_{\max}$ denote the maximum permissible ground reaction force and torque, respectively.
% These constraints can be reformulated into the compact form: ${\boldsymbol{c}}_i^- \leq {\boldsymbol{C}}_i {\boldsymbol{u}}_i \leq {\boldsymbol{c}}_i^+$
At each sampling interval, the MPC optimization problem is constructed as follows:
\begin{equation}
\begin{aligned}
& \min_{\boldsymbol{x},\boldsymbol{u}} \sum_{k=0}^{N-1} \Big[ \|\boldsymbol{x}_{k+1} - \boldsymbol{x}_{k+1,\text{ref}}\|_{\boldsymbol{Q}_k}^2 + \|\boldsymbol{u}_k\|_{\boldsymbol{R}_k}^2 \Big] \\
& \text{s.t.} \quad \boldsymbol{x}_{k + 1} = \hat{\boldsymbol{A}}_k \boldsymbol{x}_k + \hat{\boldsymbol{B}}_k \boldsymbol{u}_k, \quad k = 0, \ldots, N - 1 \\
& \phantom{\text{s.t.}} \quad \boldsymbol{c}_k^- \leq \boldsymbol{C}_k \boldsymbol{u}_k \leq \boldsymbol{c}_k^+, \quad k = 0, \ldots, N - 1 ,
\label{eq:mpc}
\end{aligned}
\end{equation}
where $N$ is the horizon. ${\boldsymbol{Q}}_k$ and ${\boldsymbol{R}}_k$ are diagonal positive definite weight matrices. The continuous SRB dynamics in Eq. \ref{eq:srbd} are discretized as described in \cite{li2021force}, yielding the system matrices ${\boldsymbol{\hat{A}}_k}$ and ${\boldsymbol{\hat{B}}_k}$. The matrices ${\boldsymbol{C}}_k$ and vectors ${\boldsymbol{c}}_k^+$ and ${\boldsymbol{c}}_k^-$ correspond to the inequality constraints defined in Eq. \ref{eq:constrain}.

% 跟踪轨迹没写清楚！！！！！！！！！！

\subsection{Soccer Dynamics and Motion Prediction}
Accurately predicting soccer ball motion requires a well-defined dynamic model. For simplicity, we approximate the ball as a point mass and neglect rotational effects. When the ball is airborne, its motion is modeled using free-fall dynamics. To describe the collision dynamics between the robot's foot and the ball, we employ an impulse-based collision model. During the instantaneous collision process, the normal direction dynamics are governed by the classical coefficient of restitution model \cite{Nonomura2010AnalysisOE}. The desired velocity of the robot foot is given by:  

\begin{equation}
\boldsymbol{v}_{des} = \frac{1}{1+\alpha}(\alpha\boldsymbol{v}_{in} + \boldsymbol{v}_{out}),
\label{eq:contact_vel}
\end{equation}  
where \(\boldsymbol{v}_{des}\) is the desired foot velocity at impact, and \(\alpha \in (0,1)\) is the restitution coefficient. \(\boldsymbol{v}_{in}\) and \(\boldsymbol{v}_{out}\) represent the ball's velocities before and after the collision, respectively.

\subsection{Swing Foot Trajectory Planning}

As shown in Fig. \ref{fig:kick}(a), the swing foot trajectory planning for both regular walking (1) (2) and kicking tasks  (3) (4) follows the same goal: the foot moves smoothly from the current position (Point A) to an intermediate point (Point B) and then to the target foothold (Point C).  
For clarity, we define the foot state, \(\boldsymbol{s}_f\), which includes the position \(\boldsymbol{p}_f\), velocity \(\boldsymbol{v}_f\), and acceleration \(\boldsymbol{a}_f\):
\begin{equation} 
\boldsymbol{s}_f = \left[ \begin{array}{c} \boldsymbol{p}_f \ \boldsymbol{v}_f \ \boldsymbol{a}_f \end{array} \right] \in \mathbb{R}^{9}.
\end{equation}

The initial foot state \(\boldsymbol{s}_f^{init}\) at point A is obtained through state estimation and kinematics calculation.  
For the kicking task, the target state \(\boldsymbol{s}_f^{kick}\) at point B is set using an RViz-based aiming tool \cite{kam2015rviz}. In regular walking, point B is simply the symmetric midpoint \(\boldsymbol{s}_f^{walk}\) between the start and end points. Both cases are unified as the intermediate state \(\boldsymbol{s}_f^{inter}\).
The target foothold state \(\boldsymbol{s}_f^{hold}\) at point C is calculated using a heuristic foot placement strategy \cite{raibert1986legged}.

\begin{equation}  
{ \boldsymbol{p}}_{f,i}^{hold} = {{ \boldsymbol{p}}} + {{\boldsymbol{v}}}\Delta t/2 + k_f \left({{{{\boldsymbol{v}}}} - {\boldsymbol{v}}^{des}}\right),  
\end{equation}  
where \(\Delta t\) is the gait cycle period, ${\boldsymbol{v}}^{des}$ is the desired of robot CoM , and $k_f$ is the proportional factor.

The STOFT planner $\boldsymbol{\mathcal{S}(\cdot)}$ generates feasible foot trajectories $\boldsymbol{\phi}(t)$ by solving a spatio-temporal optimization problem with state constraints at key points (A, B, C), which is detailed in the section \ref{cha:traj}.
\begin{equation}  
\begin{aligned}
\boldsymbol{\phi}(t)=\boldsymbol{\mathcal{S}}(\boldsymbol{s}_f^{init} ,\boldsymbol{s}_f^{inter},\boldsymbol{s}_f^{hold}) ,
\end{aligned}
\end{equation}  
where the desired foot state \(\boldsymbol{s}_f^{des}\) is obtained in real time using the trajectory function according to real time $t$.
The swing phase duration, $T_{\text{swing}}$, is adaptively optimized by the STOFT planner. Therefore, during the kicking task, the swing phase duration must be flexibly adjusted, as detailed in the section \ref{cha:gait}.
For regular walking tasks, a fixed swing phase duration is achieved through time normalization $t_{\text{norm}} = t / T_{\text{swing}}$.

% 姿态插值
The desired foot orientation $\boldsymbol{R}_{\text{des}}$ is interpolated using the rotation matrix approach:  

\begin{equation}  
\begin{aligned}
\boldsymbol{R}_{\text{des}} = \boldsymbol{R}_{\text{init}} \exp(t/T_l \cdot \log(\boldsymbol{R}_{\text{init}}^T \boldsymbol{R}_{\text{end}}) ),
\end{aligned}
\end{equation}  
where \(\boldsymbol{R}_{\text{init}}\) and \(\boldsymbol{R}_{\text{end}}\) are the initial and target foot orientations, respectively. $T_l$ is the $l$-th trajectory duration.

\subsection{Adaptive Gait Schedule}  \label{cha:gait}
To enhance the robot's walking stability and prevent issues such as foot-dragging or missteps, a double support phase is introduced. Specifically, a stance-hold phase is added between the transition from the stance to the swing phase. To manage this adaptive gait schedule, a Finite State Machine (FSM) is designed for each leg, detailing specific state definitions and transition rules.

\begin{algorithm}[H]
\caption{Leg State Machine Transition Rules}
\begin{algorithmic}[1]
\State \textbf{State Definitions:}
\State $\text{SWING}$: Swing phase, the leg is in the air.
\State $\text{STANCE}$: Stance phase, the leg need to supports the robot.
\State $\text{STANCE\_HOLD}$: Stance-hold phase, the leg maintains support while waiting for transition conditions.

\State \textbf{Transition Rules:}
\If{Current state is $\text{SWING}$}
    \If{Swing phase duration exceeds its threshold \textbf{and} kick football is finished}
        \State Transition to $\text{STANCE}$ state
    \EndIf
\ElsIf{Current state is $\text{STANCE}$}
    \If{Stance phase duration exceeds its threshold}
        \State Transition to $\text{STANCE\_HOLD}$ state
    \EndIf
\ElsIf{Current state is $\text{STANCE\_HOLD}$}
    \If{Stance-hold phase duration exceeds its threshold \textbf{and} no other leg is in $\text{SWING}$ state}
        \State Transition to $\text{SWING}$ state
    \EndIf
\EndIf
\end{algorithmic}
\end{algorithm}

\begin{figure}[h]
\centering
\includegraphics[width=\linewidth]{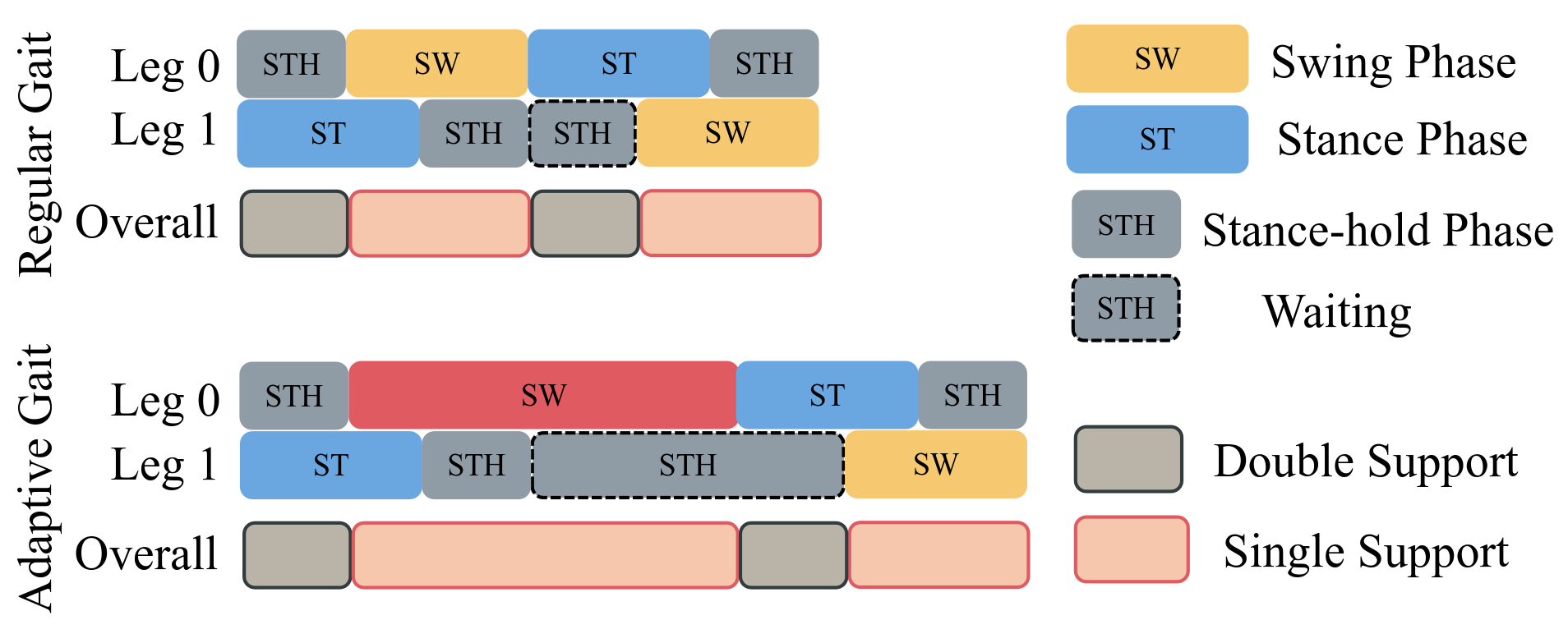}
\caption{Gait Schedule. The upper part shows the regular gait. The lower section shows the adaptive gait during kicking, where red boxes mark the swing phase extended by the STOFT planner compared to regular swing.
The first two rows indicate the FSM states of the legs, and the third row shows the overall gait.
The dashed box means the stance-hold phase has exceeded the threshold and is waiting for the other leg to exit the swing phase.
"Single support" and "double support" denote whether one or both legs support the robot.
% 这个给出两种情况，上面的图是regular gait，下面是有踢球任务时候的adaptive gait，每种情况的前两行是两条腿的fsm状态，第三行是整体的表现
% 虚线框表示Stance-hold phase duration 已经exceeds the threshold正在等待另外一条腿切换出swing phase. Single support 和 double support 分别表示 单条支撑腿，和两条支撑腿。
}
\label{fig:schedule}
\end{figure}

As illustrated in Figure \ref{fig:schedule}, during normal walking, a regular gait timetable is generated. In addition, the schedule is adaptable to scenarios that require an extended swing phase, such as when the robot performs a kicking motion.

\subsection{Torque Calculation}  

For the swing leg, the Cartesian PD controller calculates the force and torque of the \(i\)-th swing foot as follows:  
\begin{equation}  
\begin{aligned}  
\boldsymbol{f}_{swing,i} &= \boldsymbol{K}_p (\boldsymbol{p}_{f,i}^{des} - \boldsymbol{p}_{f,i}) + \boldsymbol{K}_d (\boldsymbol{\dot{p}}_{f,i}^{des} - \boldsymbol{\dot{p}}_{f,i}) \\  
\boldsymbol{\tau}_{swing,i} &= \boldsymbol{K}_R \text{Log}(\boldsymbol{R}_{f,i}^{des} \boldsymbol{R}_{f,i}^T)_\theta + \boldsymbol{K}_d (\boldsymbol{\omega}_{f,i}^{des} - \boldsymbol{\omega}_{f,i}) \\  
\boldsymbol{u}_{i} &= \begin{bmatrix} \boldsymbol{f}_{swing,i}  \ \boldsymbol{\tau}_{swing,i} \end{bmatrix}  .
\end{aligned}  
\end{equation}  

For the stance leg, the desired wrench \(\boldsymbol{u}_i\) is computed using the MPC formulation (Eq. \ref{eq:mpc}). The corresponding joint torques are mapped through the Jacobian matrix \(\boldsymbol{J}_{f,i}\) of $i$-th leg as:  
\begin{equation}  
\boldsymbol{\tau}_{joint,i} = \boldsymbol{J}_{f,i}^T \boldsymbol{u}_i  .
\end{equation}

\section{ Spatial-temporal Trajectory Optimization}  \label{cha:traj}
In this section, we present the proposed STOFT Planner, which utilizes the MINCO (MINimum COntrol) polynomial trajectory framework \cite{Wang_2022}. This method enables spatial-temporal decoupled optimization while ensuring dynamic feasibility and adhering to self-collision constraints.

\subsection{Trajectory Definition}
For foot trajectory planning, we focus on the 3-dimensional position, with the orientation interpolated on the Lie group. To ensure the trajectory passes through the intermediate point \(s_{inter}\) (which could be either \(s_{kick}\) or \(s_{walk}\)), we design a $M=2$-segment trajectory, as illustrated in Fig. \ref{fig:kick}.  
In the first segment, depicted as (1) and (3) in Fig. \ref{fig:kick}(a), the trajectory begins at the current foot state \(s_{cur}\) and reaches the state at the intermediate point \(s_{inter}\). 
% The second segment, represented by (2) and (4) in Fig. \ref{fig:kick}(a), starts from the state at \(s_{inter}\) and reaches to the final foot holding position \(s_{hold}\).
% In the first segment, shown as (1) and (3) in Fig. \ref{fig:kick}(a), the trajectory starts at \(s_{cur}\) and reaches \(s_{inter}\).  
The second segment, represented by (2) and (4), continues from \(s_{inter}\) to the final state \(s_{hold}\).  
Each segment is represented by 3-dimensional 5-order polynomial:

\begin{equation}
\boldsymbol{\phi}_l(t) = \boldsymbol{c}_l^T \boldsymbol{\beta}_l(t), \quad \boldsymbol{\beta}_l(t) = [1,\, t,\, \ldots,\, t^{5}], \ t \in [0,T_l],
\end{equation}
where $\boldsymbol{c}_l \in \mathbb{R}^{6 \times 3}$ is the coefficient matrix,
$\boldsymbol{\beta}_l(t)$ is natural basis vector, and $T_l$ is the duration of the piece of the i-th segment.

Then, using the MINCO, the trajectory \(\boldsymbol{\phi}(t)\) is parameterized by the time durations of each segment \(\boldsymbol{T}=[T_1, \ldots, T_{M}]\) and the intermediate waypoints \(\boldsymbol{q}=[q_1, \ldots, q_{M-1}]\) through a convenient space-time deformation \(\mathcal{M}\).

\begin{equation} \boldsymbol {c} = \mathcal {M}(\boldsymbol {q}, \boldsymbol {T}).  \end{equation}

%------------------------------实验部分------------------------------
\begin{figure*}[h]
  \centering
  % ========== 第一组：Toe Kick ==========
  \parbox[c]{0.05\textwidth}{
    \rotatebox{90}{\text{Toe Kick}}
  }
  \subfigure{
    \parbox[c]{0.20\textwidth}{
      \includegraphics[width=0.18\textwidth]{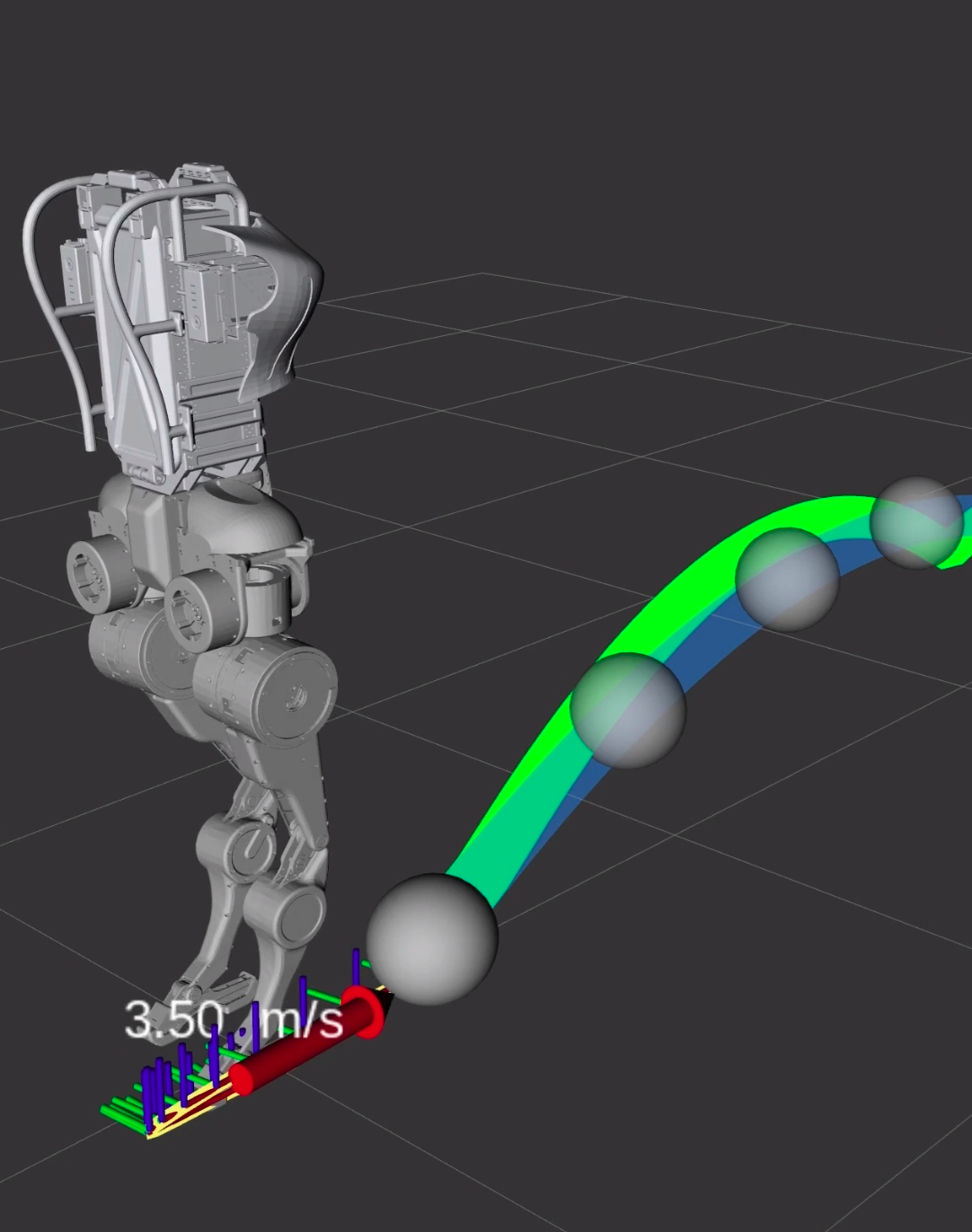}
    }
  }
  \subfigure{
    \parbox[c]{0.65\textwidth}{
      \includegraphics[width=0.65\textwidth]{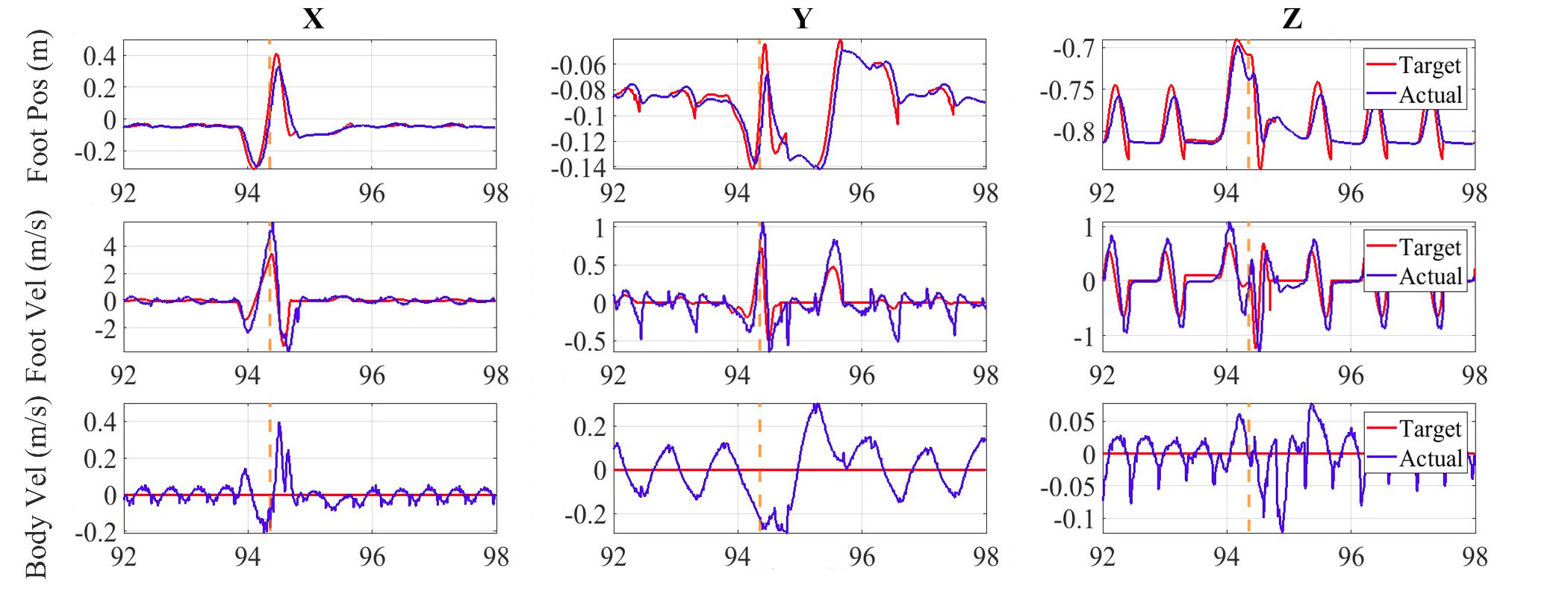}
    }
  }
  
  % ========== 第二组：Inside Foot Kick ==========
  \parbox[c]{0.05\textwidth}{
    \rotatebox{90}{\text{Inside Foot Kick}}
  }
  \setcounter{subfigure}{0}
  \subfigure[Snap shots]{
    \parbox[c]{0.2\textwidth}{
      \includegraphics[width=0.18\textwidth]{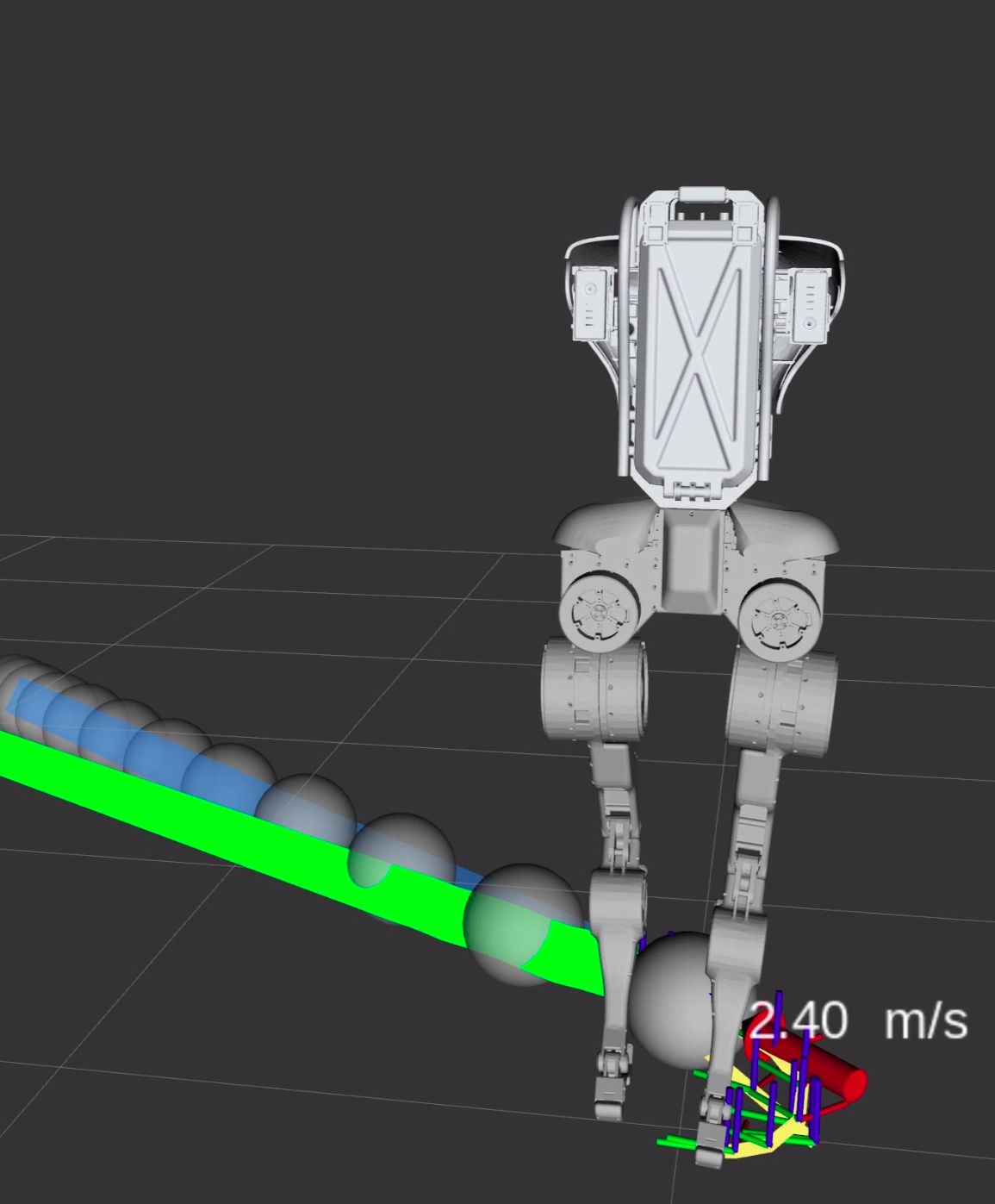}
    }
  }
  \subfigure[Data]{
    \parbox[c]{0.65\textwidth}{
      \includegraphics[width=0.65\textwidth]{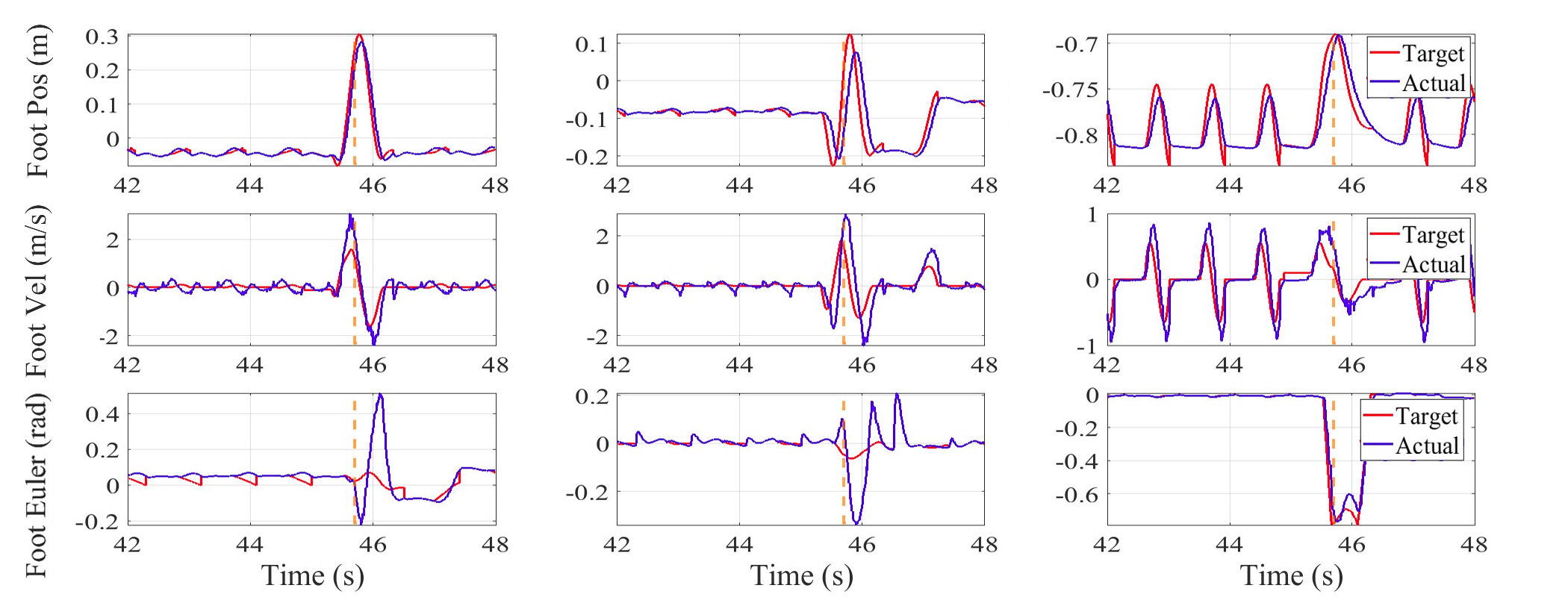}
    }
  }
  
  \caption{
    Two sets of simulation experiments. The first set features the toe kick, while the second set demonstrates the robot's inside foot kick. 
    In part (a), the snapshots illustrate the experiments, where the blue trajectory represents the predicted path, and the green trajectory indicates the actual path.
    Part (b) presents the experimental data in three rows: the top row compares the foot position reference with the actual data in the hip frame, the middle row displays foot velocity tracking in the hip frame and the bottom row showcases the robot's body velocity along with the foot Euler angles for both the toe kick and inside foot kick experiments, respectively. The orange dashed line indicates the timing of the hit.
  }
  \label{exp:simKicking}
\end{figure*}

\subsection{Problem Formulation}
The foot position planning can be formulated as an optimization problem:
\begin{equation}
\begin{split}
\min_{\boldsymbol{q},\,\boldsymbol{T}} \sum_{l=0}^{M} \Bigl(\int_{0}^{T_l} ||\boldsymbol{\phi}^{(3)}_l(t)||^{2} \ \mathrm{d}t +\rho  T_l\Bigr) \\
+\sum\lambda_{\star}\mathcal{I}_{\star}(\boldsymbol {c}(\boldsymbol {q}, \boldsymbol {T}), \boldsymbol {T}) ,
\end{split}
\label{eq:obj_func}
\end{equation}

\text{subject to:}
\begin{subequations}\label{eq:all}
\begin{align} 
  T_l &> 0, \quad \forall l \in \{0, 1\}, \label{eq:Ti}\\
  % 起点约束
  \boldsymbol{\phi}^{(0)}_0(0) &= \boldsymbol{p}_f^{init}, \label{eq:phi_start}\\
  \boldsymbol{\phi}^{(1)}_0(0) &= \boldsymbol{v}_f^{init}, \label{eq:phi_start_v}\\
  % 终点约束
  \boldsymbol{\phi}^{(0)}_1(T_1) &= \boldsymbol{p}_f^{hold}, \label{eq:phi_end}\\
  \boldsymbol{\phi}^{(1)}_1(T_1) &= \boldsymbol{v}_f^{hold}, \label{eq:phi_end_v}\\
  % 碰撞点连续性
  \boldsymbol{\phi}^{(0)}_0(T_0) &= \boldsymbol{\phi}^{(0)}_1(0) = \boldsymbol{p}^{inter}_f, \label{eq:phi_con}\\
\boldsymbol{\phi}^{(1)}_0(T_0) &= \boldsymbol{\phi}^{(1)}_1(0) = \boldsymbol{v}^{inter}_f, \label{eq:phi_con_v}\\
\end{align}
\end{subequations}
% 这里缺少了对J的说明，之后cost合并到一起
where $\mathcal{I}_{\star}$ is the generalized time integral penalty
with generalized weight $\lambda_{\star}$, including smoothness, time, dynamical feasibility, safety constraints, respectively, which will be detailed in the following sections.
The subscript $\star$ serves as a placeholder for specific function types.
The boundary condition \eqref{eq:Ti} ensures that each time segment $T_i$ remains positive, while the formulations in \eqref{eq:phi_start}, \eqref{eq:phi_start_v}, \eqref{eq:phi_con}, \eqref{eq:phi_con_v}, \eqref{eq:phi_end} and \eqref{eq:phi_end_v} define the initial, intermediate, and final conditions of the polynomial trajectory.
We use different types of cost functions $\mathcal{G}_{\star}$ to achieve those constraints. 

\begin{equation}
\mathcal{I}_{\star} = \sum_{l=1}^M \frac{T_l}{\kappa} \sum_{j=0}^{\kappa_l} \bar{\omega}_j \mathcal{G}_{\star}(c_l, \frac{j}{\kappa}T_l),
\end{equation}
where $\kappa$ denotes the number of samples on a trajectory segment, and $\frac{j}{\kappa}$ represents the normalized timestamp. $(\bar{\omega}_0, \bar{\omega}_1,..., \bar{\omega}_{\kappa_l-1}, \bar{\omega}_{\kappa_l}) = (1/2, 1,..., 1, 1/2)$ are the quadrature coefficients.
Eq. \ref{eq:obj_func} can be expressed as $J(\boldsymbol{q},\boldsymbol{T})$ for simplicity.
To solve the optimization problem, it is necessary to compute the gradients $\frac{\partial J(\boldsymbol{q}, \boldsymbol{T})}{\partial \boldsymbol{q}}$ and $\frac{\partial J(\boldsymbol{q}, \boldsymbol{T})}{\partial \boldsymbol{T}}$
, which can be derived using the chain rule:
\begin{equation}
\frac{\partial J(\boldsymbol{q}, \boldsymbol{T})}{\partial \boldsymbol{q}} = \frac{\partial J}{\partial \boldsymbol{c}} \cdot \frac{\partial \boldsymbol{c}}{\partial \boldsymbol{q}}, \quad
\frac{\partial J(\boldsymbol{q}, \boldsymbol{T})}{\partial \boldsymbol{T}} = \frac{\partial J}{\partial \boldsymbol{T}} + \frac{\partial J}{\partial \boldsymbol{c}} \cdot \frac{\partial \boldsymbol{c}}{\partial \boldsymbol{T}}.
\label{eq:grad}
\end{equation}

The cost function can be constructed in a similar manner for all subsequent constraints:
\begin{equation} \mathcal{I}_{\star} = \max{\left[\mathcal{G}_{\star}^{3}, \mathbf{0}\right] }. \end{equation}

\subsection{Time penalty and duration constraints}
To ensure that the swing phase duration within a feasible range. We introduce a cubic penalty function to enforce bounds on each segment duration $T_l$:

\begin{equation}
\mathcal{G}_t(T_l) = 
  \begin{cases} 
  0, & \text{if }  T_l \leq T_{\text{max}} \\
  (T_l - T_{bound})^3, & \text{otherwise},
  \end{cases}
\end{equation}
where $T_{\text{max}}$ is the max swing phase duration.

\subsection{Dynamical feasibility constraints}

Considering the maximum allowable velocity $v_{\text{max}}$ and acceleration $a_{\text{max}}$ of the foot, we define the dynamical feasibility constraints as:

\begin{equation}
  \mathcal{G}_f(\boldsymbol{\xi}) = 
  \begin{cases} 
  0, & \text{if } \|\boldsymbol{\xi}\| \leq \xi_{\text{max}} \\
  \|\boldsymbol{\xi}\|^2 - \xi_{\text{max}}^2, & \text{if } \|\boldsymbol{\xi}\| > \xi_{\text{max}},
  \end{cases}
\end{equation}
where $\boldsymbol{\xi}$ represents the trajectory velocity $\boldsymbol{v}$ or acceleration $\boldsymbol{a}$ at a given time point, and $\xi_{\text{max}}$ represents the corresponding constraint limit $v_{\text{max}}$ or $a_{\text{max}}$.

\subsection{Self collision avoidance}
Generating self-collision-free trajectories is crucial for ensuring successful kicking motions. By approximating the foot as a sphere with radius $\epsilon_r$, we define an Euclidean Signed Distance Field (ESDF) $\mathcal{D}(\boldsymbol{p}_{any})$ to represent the distance from any point $\boldsymbol{p}_{any}$ to the nearest obstacle $\boldsymbol{p}_{obs}$:

\begin{equation}
\mathcal{D}(\boldsymbol{p}_{any}) = 
\begin{cases} 
-\min_{\boldsymbol{q} \in \mathcal{O}}\|\boldsymbol{p}_{any} - \boldsymbol{p}_{obs}\|, & \text{if } p \text{ \ inside obstacle} \\
\min_{\boldsymbol{q} \in \mathcal{O}}\|\boldsymbol{p}_{any} - \boldsymbol{p}_{obs}\|, & \text{otherwise},
\end{cases}
\end{equation}
where $\mathcal{O}$ represents the set of occupied spaces. To achieve self-collision avoidance, we dynamically update the ESDF, treating other leg as obstacles. 

Based on this, we define the collision-free constraint as:

\begin{equation}
\mathcal{G}_c(\boldsymbol{p}_{any}) = 
  \begin{cases} 
  0, & \text{if } \mathcal{D}(\boldsymbol{p}_{any}) - \epsilon_r \geq 0 \\
  \epsilon_r - \mathcal{D}(\boldsymbol{p}_{any}), & \text{if } \mathcal{D}(\boldsymbol{p}_{any}) - \epsilon_r < 0,
  \end{cases}
\end{equation}
where $\epsilon_r$ is the radius of the sphere representing the foot. This constraint ensures that the foot maintains a minimum safe distance from all obstacles, effectively preventing collision.

\subsection{Bipedal robot kinematic constraints}
To ensure that the generated trajectories adhere to the physical constraints of the robot's leg length, we constrain the foot position within a spherical workspace centered at the robot's hip position $ \boldsymbol{p}_{hip}$ with radius $\epsilon_h$:

\begin{equation}
\mathcal{G}_h(\boldsymbol{p}_{any}) = 
  \begin{cases} 
  0, & \text{if } \|\boldsymbol{p}_{any} - \boldsymbol{p}_{hip}\| \leq \epsilon_h \\
  \|\boldsymbol{p}_{any} - \boldsymbol{p}_{hip}\| - \epsilon_h, & \text{if } \|\boldsymbol{p}_{any} - \boldsymbol{p}_{hip}\| > \epsilon_h.
  \end{cases}
\end{equation}

In summary, the complete cost function presented in Eq. \ref{eq:obj_func} has been formulated. By leveraging this cost function together with the gradient computation method shown in Eq. \ref{eq:grad}, we can efficiently solve the optimization problem to generate the optimal foot trajectory. The optimization problem is solved using the L-BFGS algorithm, as implemented in the open-source library presented by \cite{Wang_2022}.

\begin{figure}[h]
    \centering
    \includegraphics[width=0.98\linewidth]{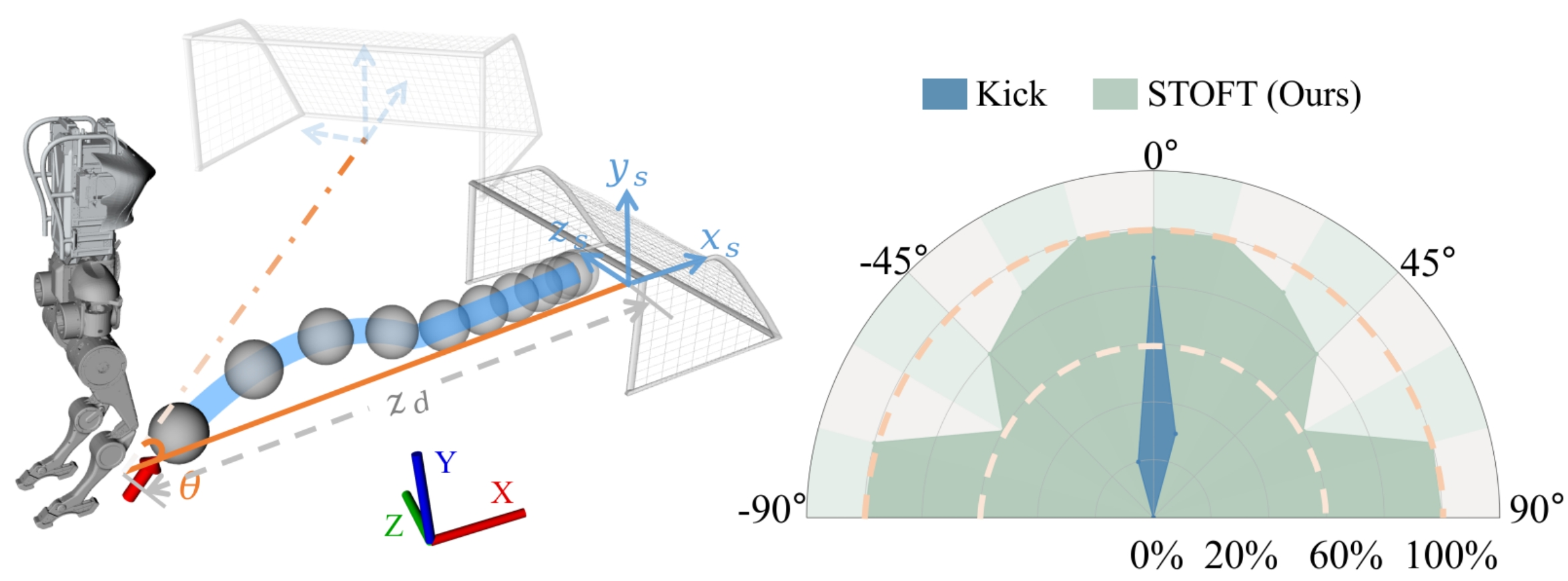}
       \caption{Accuracy test for the STOFT. The left figure illustrates the testing scenario where kicking angles $\theta$ vary from -90° to 90° with the robot positioned $z_d = 3 \ m$ from the soccer goal, while the right figure presents the success rates of both the STOFT planner and the baseline methods across different angles. The STOFT planner consistently achieves a higher scoring success rate than the baseline methods.}
    \label{fig:accuracy}
\end{figure}

\begin{figure}[h]
    \centering
    
    % First subfigure with label (a)
    \subfigure[Foot Trajectory]{
        \parbox[c]{\linewidth}{
            \centering  % 添加居中控制
            \includegraphics[width=0.9\linewidth]{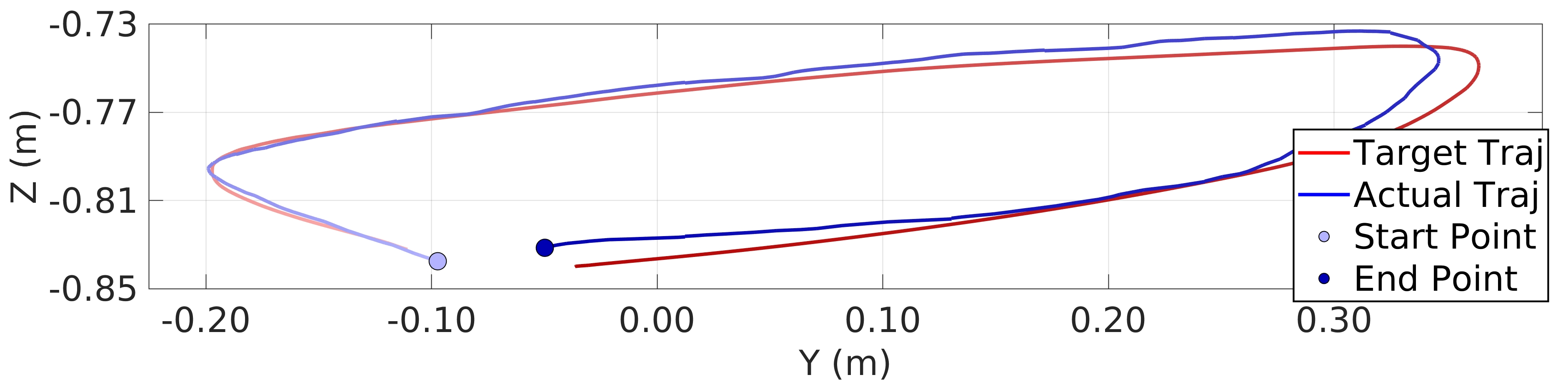}
        }
        \label{fig:traj}
    }
    
    % Second subfigure with label (b)
    \subfigure[Gait Phase]{
        \parbox[c]{\linewidth}{
            \centering  % 添加居中控制
            \includegraphics[width=0.9\linewidth]{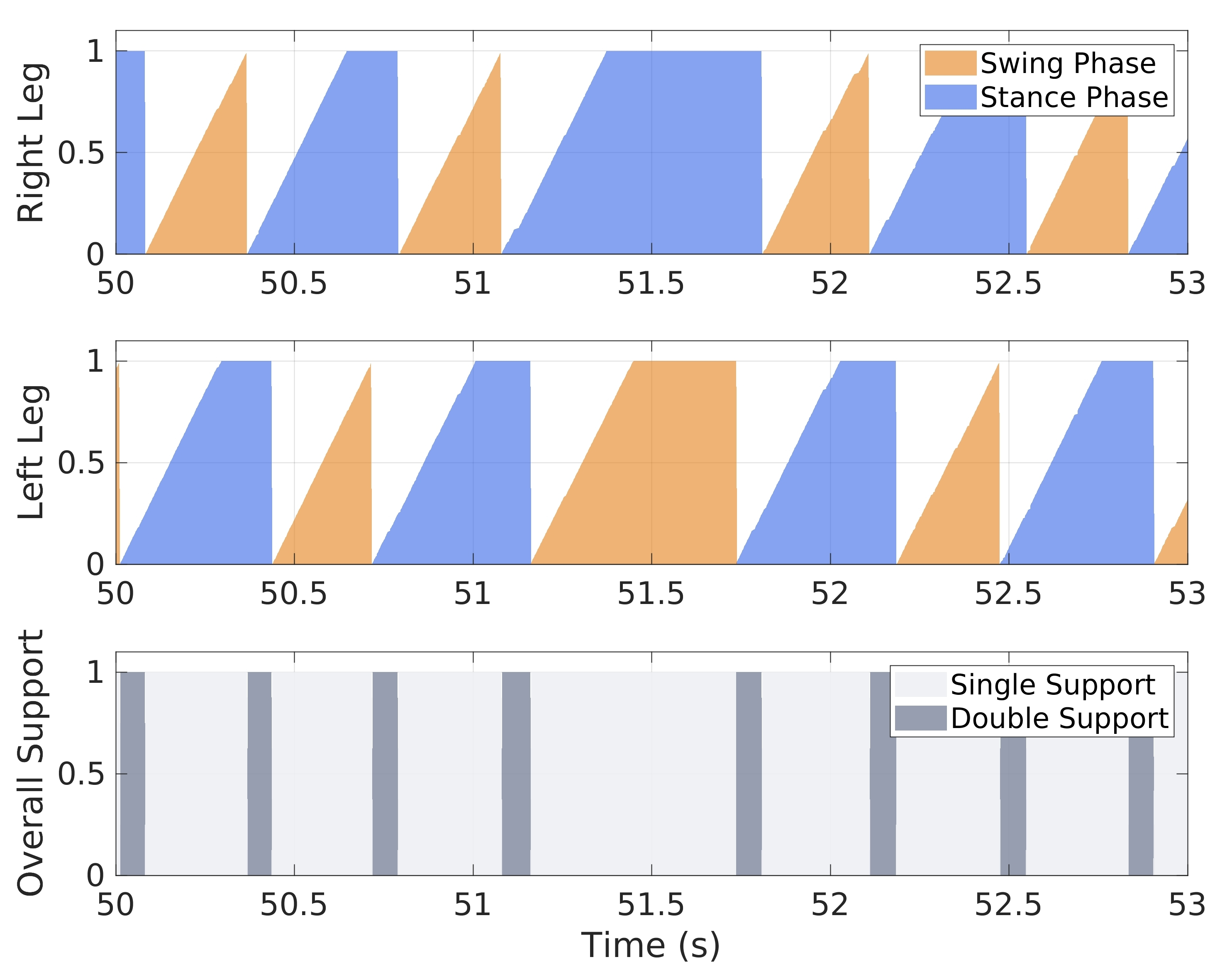}
        }
        \label{fig:gait}
    }
    
    \caption{
        (a) demonstrates the target and actual foot trajectories in the body frame when executing a 3 m/s kicking motion. 
        (b) shows the corresponding gait schedule, where the first and second rows represent the gait schedules for the right and left legs, respectively. 
        Orange blocks indicate the stance phase, while blue blocks denote the swing phase, with phase counters ranging from 0 to 1. 
        The third row displays the overall gait, where light gray represents the single support phase and dark gray indicates the double support phase.
    }
    \label{fig:trajandgait}
\end{figure}
  
\section{Experimental Results}

\subsection{Implementation Details}

We conducted comprehensive experiments on both simulation and hardware platforms to evaluate the system's performance, including toe kicks, inside foot kicks, shooting accuracy, foot trajectory, and gait schedule. Detailed analyses of these results are provided in the following sections.  

For simulation experiments, we employed the MuJoCo physics engine \cite{todorov2012mujoco}, with the simulated robot model configured to match the physical parameters of its real-world counterpart, including properly tuned collision dynamics. All simulations were executed on a desktop computer equipped with an Intel Core i7-12700H CPU and 32 GB of RAM. For dynamic constraints, we applied the following parameter limits: $v_{\max}=5$ m/s, $a_{\max}=15$ m/s$^2$, $T_{\max}=1.2$ s, with the collision coefficient $\alpha$ determined through offline calibration to be approximately 0.65.

To validate the feasibility and performance of the STOFT planner in shooting accuracy, we established a controlled test scenario using a soccer goal with dimensions $0.9 m \times 0.6 m \times 0.6 m$ and a regulation size football.  
For hardware experiments, we recorded foot trajectory performance and gait schedule, demonstrating the STOFT planner's effectiveness in real-world applications.  

\subsection{Toe Kick Experiments}
As shown in Fig. \ref{exp:simKicking}, the toe kick experiment involved instructing the robot to kick a ball positioned 0.3 meters ahead at a target velocity of 3.5 m/s. At approximately 94.5 seconds (see Fig. 3), the ball was struck. Although the robot's foot continued to move slightly forward, it had already begun to decelerate. The kick introduced a transient disturbance in the robot’s body velocity, but the system quickly restored stability.

\subsection{Inside Foot Kick Experiments}
For scenarios requiring a larger kicking angle, the Inside Foot Kick strategy was adopted. The robot was instructed to kick a soccer ball positioned to its front-left at a target speed of 2.4 m/s. As shown in Fig. \ref{exp:simKicking}, the third row illustrates the foot's orientation relative to the hip frame. The roll angle is uncontrollable and control efforts are primarily focused on the yaw axis. Upon ball impact, a noticeable yaw deviation occurs; however, the foot quickly recovers and catches up with its desired trajectory.

\subsection{Accuracy Test for STOFT}

For a comprehensive assessment, we designed kicking tasks with varying angles ranging from -90° to 90°, with the soccer ball placed 3 meters from the robot and aimed at the goal. Each scenario was tested ten times to ensure statistical significance, and the overall success rate (SR) was recorded.

Our baseline reference is the 2024 RoboCup champion team, UCLA's RoMeLa lab, which employed their ARTEMIS robot. Their strategy involves a straightforward forward kick, which is highly effective when the ball is directly in front of the robot. However, when the ball is positioned off center, the success rate drops significantly as the robot struggles to accurately redirect the ball.

In contrast, our proposed method takes a flexible approach.  When the ball is directly in front of the robot, we use the "toe kick" method, allowing precise control over the foot's position and impact speed, resulting in a nearly 100 \% success rate. When the ball is positioned to the side, we employ the "inside foot kick" strategy. Although this approach might seem counterintuitive, it also achieves a 100 \% success rate. The key is that the inside foot kick requires minimal foot orientation adjustment—simply swinging the leg sideways is sufficient to accurately score the goal.
As illustrated in Fig. \ref{fig:accuracy}, the proposed STOFT planner consistently outperforms the baseline method in terms of scoring success rate. For angles between -30° and 30°, as well as 60° to 90°, the STOFT planner achieves a perfect 100 \% success rate. When handling angles from -90° to -60°, we utilize the left foot for kicking, maintaining the same high success rate.

\subsection{Foot Trajectory and Gait Schedule}
In hardware experiments, fig. \ref{fig:traj} presents the foot-end trajectory when executing a kicking motion that achieves 3 m/s velocity at the ball impact point (located 0.2 m ahead of the robot). 
The ball collision induces limited disturbance to the foot-end trajectory.
The corresponding gait schedule, shown in Fig. \ref{fig:gait}, indicates that during regular walking, the robot maintained a periodic gait pattern. 
 At t $=$ 51.2 s, the left leg initiated a kicking motion lasting 0.59 s, during which the right stance phase was adaptively extended to accommodate the dynamic demands of the kick.
%------------------------------总结------------------------------
\section{CONCLUSION}

We present an integrated system for dynamic bipedal robot shooting, combining vision-based ball detection, interactive aiming, robot state estimation, MPC balance control, STOFT planner, and adaptive gait generation. 
Our key contribution, the STOFT Planner, employs spatial-temporal optimization to solve kicking motion planning in under 1 ms while satisfying dynamic constraints and avoiding self-collisions, generating human-like backswing motions for precise execution.
Experimental validation using the PEARL humanoid robot demonstrates competition-ready performance with reliable, high-precision shooting capabilities.

%------------------------参考文献------------------------
\bibliographystyle{IEEEtran}%引用类型
% \bibliography{Mybib}%引用文件列表

\end{document}